# Handling Boundary Constraints for Numerical Optimization by Particle Swarm Flying in Periodic Search Space

Wen-Jun Zhang
Institute of Microelectronics,
Tsinghua University,
Beijing 100084, P. R. China
Email: zwj@tsinghua.edu.cn

Xiao-Feng Xie
Institute of Microelectronics,
Tsinghua University,
Beijing 100084, P. R. China
Email: xiexf@ieee.org

De-Chun Bi
Department of Environmental Engineering,
Liaoning University of Petroleum
& Chemical Technology
Fushun, Liaoning, 113008, P. R. China

***Abstract-*** **The *Periodic* mode is analyzed together with two conventional boundary handling modes for particle swarm. By providing an infinite space that comprises periodic copies of original search space, it avoids possible disorganizing of particle swarm that is induced by the undesired mutations at the boundary. The results on benchmark functions show that particle swarm with *Periodic* mode is capable of improving the search performance significantly, by compared with that of conventional modes and other algorithms.**

## I. Introduction

The numerical optimization problems (NOP) can be defined as finding $\vec{x} \in S \subseteq \mathbb{R}^D$ such that

$$\begin{cases} f(\vec{x}) = \min\{f(\vec{y}); \vec{y} \in S\}, \\ g_j(\vec{x}) \le 0, \quad for\ j \in [1,m] \end{cases} \quad (1)$$

where $\vec{x} = (x_1,...,x_d,...,x_D)$ ( $1 \le d \le D$ , $d \in \mathbb{Z}$ ), and $x_d \in [l_d, u_d]$, $l_d$ and $u_d$ are lower and upper values of the boundary constraints, respectively. $f$ and $g_j$ are functions on $S$; $S$ is a $D$-dimensional space defined as a Cartesian product of domains of variables $x_d$'s. The set of points, which satisfying all the constraint functions $g_j$, is denoted as feasible space ($S_F$).

Particle swarm optimization (PSO) [6, 11] is a novel stochastic algorithm inspired by social behavior of swarms. Each agent, call *particle*, flies in a $D$-dimensional space $S$ according to the historical experiences of its own and its colleagues. The location of the $i$th ( $1 \le i \le N$, $i \in \mathbb{Z}$ ) particle is represented as $\vec{x}_i = (x_{i1},\dots, x_{id}, \dots, x_{iD})$. The best previous location of the $i$th particle is recorded and represented as $\vec{p}_i = (p_{i1},\dots, p_{id}, \dots, p_{iD})$, which is also called *pbest*. The index of the best *pbest* among all the particles is represented by the symbol $g$. The location $\vec{p}_g$ is also called *gbest*. The velocity for the $i$th particle is represented as $\vec{v}_i = (v_{i1},\dots, v_{id}, \dots, v_{iD})$. At each time step, the $i$th particle is manipulated according to the following equations for its $d$th dimension [12]:

$$\begin{aligned} v_{id} &= w \cdot v_{id} + c_1 \cdot U_{\mathbb{R}}() \cdot (p_{id} - x_{id}) + c_2 \cdot U_{\mathbb{R}}() \cdot (p_{gd} - x_{id}) \\ x_{id} &= x_{id} + v_{id} \end{aligned} \quad (2)$$

where $w$ is inertia weight, $c_1$ and $c_2$ are acceleration constants, $U_{\mathbb{R}}()$ are random real values between 0 and 1.

The particle swarm is trying to perform as a self-organizing system with extraordinarily flexibility [14]. A particle "surfs" it on waves [1, 13], which $\vec{v}$ and $\vec{x}$ are similar to kinetic and potential energy, respectively; and $\vec{p}_g$ serves as a quasi-gravity center for the swarm, especially at the end of the process.

To solving NOP, two issues should be handled: a) constraint functions and b) boundary constraints.

The constraint handling methods [2-5, 7-10] are employed for constructing the fitness landscape $F(\vec{x})$ in the $S$. To avoiding the laborious and often troublesome setting of the penalty coefficients in the methods based on penalty functions [9], and without requiring a starting point in $S_F$ [5], the handling methods that following Deb's criteria [2] is often applied [7, 15]: a) any feasible solution is preferred to any infeasible solution; b) among two feasible solutions, the one having better objective function value is preferred; c) among two infeasible solutions, the one having smaller constraint violation is preferred.

However, to satisfying boundary constraints, the conventional handling methods [7, 14], which are keeping the individuals lying inside the $S$, seem not always suitable for the efficient flying of particle swarm.

The purpose of this paper is to study a robust boundary constraints handling method for particle swarm. In the next section, the drawbacks of conventional handing methods, which keep the particles flying inside the $S$, are discussed. Then in section 3, the *Periodic* handling mode is realized by providing an infinite space for the flying of particle swarm, which is composed of periodic copies of original $S$. In section 4, experimental results in different boundary handling methods and comparison with existing results by different algorithms [3, 4, 10] are reported and discussed. In the last section, we conclude the paper.

## II. Conventional Boundary Handling

For the current swarm, it has a possible flying domain in next generation according to equations (2), as the space $S_m$ shown in Fig. 1. Its center is closed to the location of the $\vec{p}_g$, which is a gravity center of swarm.

The *gbest* is always changing its location during the evolution, which leads the swarm to moving. It means that some particles may exceed the boundary, i.e. $S_m \cap \bar{S} \ne \Phi$ (where $\Phi$ is null set), especially when the *gbest* is closer to the boundary.

The conventional boundary handling methods are trying to keep the points inside the original *S*. This means for the particles that flying into the set $S_m \cap \bar{S}$, such as $\vec{x}_M \notin S$ in Fig. 1, is invalid and should be adjusted to original $S$ by an additional mutation $\tilde{M}(\vec{x})$ appended on equations (2).

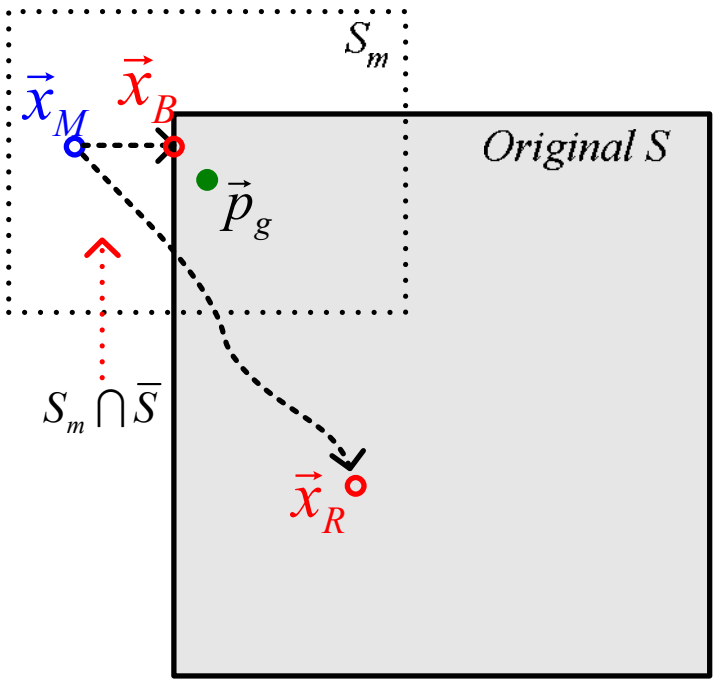

Figure 1. Schematic for conventional boundary handling modes.

Unlike other algorithms, such as genetic algorithms (GAs) [2, 3], evolution strategies [4], and differential evolution (DE) [12], etc., the status of the *t*th generation of each particle in swarm have direct effects on its status of the next generation. However, when the *gbest* is close to the boundary, the undesired mutations $\tilde{M}(\vec{x})$ may become too frequently to keep the self-organizing of swarm dynamics, which is maintained by the nonlinear interactions in swarm by equations (2).

Conventionally, there have two mainly modes are employed: *Boundary* mode and *Random* mode [7].

### *A. Boundary mode*

For the *Boundary* mode, the *d*th dimension of $\vec{x} \notin S$ is mutated to the boundary (as $\vec{x}_M \rightarrow \vec{x}_B$ in Fig. 1):

$$\tilde{M}_B(x_d): \begin{cases} x_d = l_d & \text{IF } x_d < l_d \\ x_d = u_d & \text{IF } x_d > u_d \end{cases} \quad \textbf{(3)}$$

The $\tilde{M}_B(x_{id})$ forces the particle that expects to outside the *S* returns to the boundary, its direct effects of include: a) decreasing the $v_{id}$; b) decreasing $|p_{id}\text{-}x_{id}|$ and $|p_{gd}\text{-}x_{id}|$. Both decrease the "energy" of particles, i.e. decrease the domain of $S_m$. Moreover, for worse case, the point mutated by $\tilde{M}_B$ may become the *gbest*, which may be especially occurred at the early evolution stage, since the following reasons: a) the particles would have enough "energy" values to exceed the original *S*; b) the current *gbest* is not in high fitness value. Then as a quasi-gravity center, it attracts other particles to the boundary and such particles cannot leave the boundary in the following generations unless the *gbest* leave the boundary, which accelerates the swarm into equilibrium state and may lead to the premature convergence.

### *B. Random mode*

For the *Random* mode, the *d*th dimension of $\vec{x} \notin S$ is mutated with random values (as $\vec{x}_M \rightarrow \vec{x}_R$ in Fig. 1) [7]:

$$\tilde{M}_R(x_d): x_d = U_{\mathbb{R}}(l_d, u_d) \quad \text{IF } x_d \notin [l_d, u_d] \quad \textbf{(4)}$$

where $U_{\mathbb{R}}(l_d, u_d)$ is a random value between $l_d$ and $u_d$.

The point that is mutated by $\tilde{M}_R$ has the probability $(\bar{S}_m \cap S)/S$ to exceed the $S_m$, since it is a random point in *S*. As the point exceeds the $S_m$, the direct effects of $\tilde{M}_R(x_{id})$ include: a) increasing the $v_{id}$; b) increasing $|p_{id}\text{-}x_{id}|$ and $|p_{gd}\text{-}x_{id}|$. Both effects increase the "energy" of swarm. If the global optimum is closed to the boundary, then *gbest* is closing to the boundary on if the algorithm has capability to approach the global optimum. However, such frequent $M_R$ operations keep a relative large $S_m$, which disturbs the swarm into chaos state and decrease the convergence speed to global optimum.

## III. PERIODIC SEARCH SPACE

This paper studies a new boundary handling method, which call as *Periodic* mode [15]. It provides an infinite search space for the flying of particles, which is composed of the periodic copies of original *S* with same fitness landscape, as shown in Fig. 2. Where the grey region represents the original space $S^{(O)}=S$, its neighborhood regions are its periodic copies $S^{(C)}$.

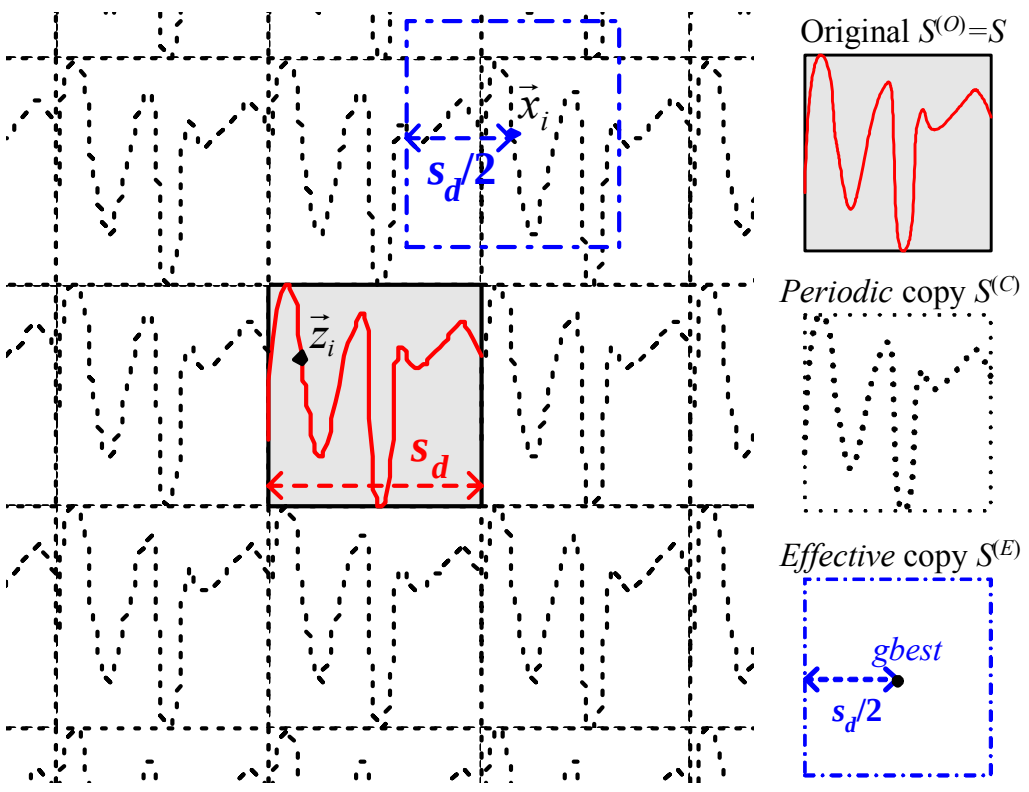

Figure 2. Schematic for *Periodic* mode.

For the *Periodic* mode, the location $\vec{x}$ of each particle is not adjusted to *S* by any mutation operations while $\vec{x} \notin S$ . However, $F(\vec{x})=F(\vec{z})$, which $\vec{z} \in S$ is the *mapping point* of $\vec{x}$ :

$$\tilde{M}_P(x_d \to z_d): \begin{cases} z_d = u_d - (l_d - x_d)\% s_d & \text{IF } x_d < l_d \\ z_d = l_d + (x_d - u_d)\% s_d & \text{IF } x_d > u_d \\ z_d = x_d & \text{IF } x_d \in [l_d, u_d] \end{cases} \quad (5)$$

where '%' is the modulus operator, $s_d = | u_d - l_d |$ is the parameter range of the $d$th dimension. The ultimate optimized point $\vec{x}^*$ is calculated from $\tilde{M}_P(gbest \to \vec{x}^*)$, which is satisfying the boundary constraints.

In *Periodic* mode, for $\forall \vec{x}$ , there exists an *effective* copy (which is called $S^{(E)}$ as in Fig. 2) of $S$, which the center point is $\vec{x}$, and for the $d$th dimension, the lower and upper boundary values are $x_{id}- s_d /2$ and $x_{id}+ s_d /2$, respectively. For each point $\vec{x}$ in $S^{(E)}$, it has a one-to-one mapping point that locating in the original $S$. To searching in $S^{(E)}$ is almost equivalent to searching in $S$.

Comparing with conventional boundary constraints handling methods, the *Periodic* mode have some advantages to enhance robustness of the particle swarm.

Firstly, to eliminate the undesired mutations caused by boundary constraints, the ratio $R(S_m, S) = (S_m \cap \bar{S})/S_m$ that outside the $S$ should be minimized, which is achieved when the center of $S_m$ (often near the *gbest*) is located near or at the center of $S$. It is possible for *Periodic* mode, since the searching space can be $S^{(E)}$ of the *gbest*.

Secondly, the performance of an algorithm is improved if the variation distance between the current *gbest* and the global optimum point is shorter, especially when the *gbest* is temporarily trapped into a local optimum at a certain stage of evolution. In fact, for $d$th dimension, the *maximum possible length* is decreased from $s_d$ (in the original $S$ for conventional handling methods) to $s_d /2$ (in the $S^{(E)}$ of *gbest* for *Periodic* mode), as shown in Fig. 2.

Of course, it should be avoid that the domain of $S_m$ being much larger than of $S^{(E)}$ of the *gbest* in some dimensions, i.e. $s_{m,d}=k \cdot s_d$ and $k$ is much larger than 1. Since at the situation, although the particle swarm still maintains the self-organization dynamics, it has less efficiency due to the many redundant states in $s_{m,d}$. Fortunately, it can be avoid as the particle swarm is convergent [1, 13], by using the following parameter settings: a) *constriction factor* [1]; and b) a small $w$ during [14] or at least after the early stage [11] of evolution.

## IV. Results and Discussion

In order to study the performance of the boundary handling methods for particle swarm, eight constrained numerical problems by Michalewicz et al. [8] and four engineering design examples [10] have been tested, which were compared with recently published results[3, 4, 10].

### *A. Michalewicz's examples* [8]

Table 1 gives the global optimum (type and value $F^*$) and the published results by EAs [3, 4] for Michalewicz's examples with inequality constraints [8]. Here for $G_1$, $G_4$, their global optimums are located at the boundary; for $G_2$, $G_6$, $G_7$, their global optimums are closed to the boundary.

TABLE 1. Global Optimum and Existing Results

| *F.* | *Type* | $F^*$ | ES [4] | GA [3] |
|---|---|---|---|---|
| $G_1$ | Min | -15 | -15.000 | -15.000 |
| $G_2$ | Max | 0.80362 | 0.56 | 0.7901 |
| $G_4$ | Min | -30665.54 | -30665.5 | -30665.2 |
| $G_6$ | Min | -6961.814 | -6961.81 | -6961.8 |
| $G_7$ | Min | 24.306 | 24.6162 | 26.580 |
| $G_8$ | Max | 0.095825 | 0.095825 | 0.095825 |
| $G_9$ | Min | 680.630 | 680.635 | 680.72 |
| $G_{10}$ | Min | 7049.248 | 7193.11 | 7627.89 |

The main parameters of $(\mu+\lambda)$-ES [4] includes: $\mu$=100, $\lambda$=300, the maximum number of generations $T$=5E3, its total evaluation times $T_E \approx \lambda \cdot T$=1.5E6.

The main parameters of GA [3] includes: the population size $N$=70, $T$=2E4, $T_E$=$N \cdot T$=1.4E6.

Two particle swarm versions were tested: a) LPS [11]: with a linear decreasing inertia weight; b) DEPS [15]: hybrid with a differential evolution (DE) operator [12].

The parameters of LPS includes: a linearly decreasing $w$ which from 0.9 to 0.4 [11], $c_1$=$c_2$=2, for $d$th dimension, the maximum velocity $v_{max,d}$=$(u_d$-$l_d)$/2.

The parameters of DEPS includes: number of particles $N$, $c_1$=$c_2$=2.05 in constriction factor [1], and for the hybrid DE operator [15], $CR$=0.9.

For both particle swarm versions, four cases were tested, as in table 2, where $T$=2E3. For all the cases, $T_E$=$N \cdot T$ are less than the cases of GA [3] and ES [4]. For each example, 100 runs were executed.

TABLE 2. Settings for Test Cases of LPS and DEPS

| Cases | #B | #R | #P1 | #P2 |
|---|---|---|---|---|
| *Mode* | *Boundary* | *Random* | *Periodic* | *Periodic* |
| *N* | 14 | 14 | 14 | 70 |
| $T_E$ | 28000 | 28000 | 28000 | 140000 |

TABLE 3. Summary of Mean Results by the Cases of LPS

| *F.* | LPS#B ($r_f$) | LPS#R ($r_f$) | LPS#P1 | LPS#P2 |
|---|---|---|---|---|
| $G_1$ | -4.998 | -1.936 (79) | -14.9140 | -14.9961 |
| $G_2$ | 0.50052 | 0.43463 | 0.71921 | 0.77867 |
| $G_4$ | -30549.87 | -30517.0 | -30665.5 | -30665.54 |
| $G_6$ | -6961.8 (96) | -4592.9 | -6961.7 | -6961.814 |
| $G_7$ | 914.790 (11) | 38.511 | 26.047 | 25.161 |
| $G_8$ | 0.095825 (2) | 0.095825 | 0.095825 | 0.095825 |
| $G_9$ | 121114.09 | 680.76 | 680.75 | 680.66 |
| $G_{10}$ | 12398 (21) | 8285.6 | 7756.6 | 7562.6 |

Table 3 gives the summary of mean best results $F_B$ by the cases of LPS. The values $r_f$ in the parenthesis gives the ratio of runs that are failed in entering $S_F$ (if there have no parenthesis, it means no failed runs) in percentage, and only those runs that are succeeded in entering $S_F$ are counted for calculating the $F_B$.

For LPS#B, LPS#R, and LPS#P1, the only difference is their boundary handling modes. It can be found that LPS#B got worse results (and a large amount of runs are even failed in entering $S_F$, especially for $G_6$, 96% runs are failed), since many runs for *Boundary* mode were premature convergence at the boundary of $S$ during the early evolution stage. For LPS#R, it could not get satisfied results in give $T$ generations due to the unnecessarily $M_R$ operations in *Random* mode, when the global optimum of a problem is located at or closed to the boundary (except for $G_8$, $G_9$, $G_{10}$), especially for $G_1$, 79% runs are failed in entering $S_F$. For LPS#P1, it performs better than LPS#B and LPS#R in almost all examples.

LPS#P2 gets better results than LPS#P1 when $N$ is increased from 14 to 70. By comparison it with the existing results by EAs in Table 2, it can be seen that LPS#P2 get worse results than ES [4], which with one better ($G_2$), three almost same ($G_4$, $G_6$, $G_8$) and four worse ($G_1$, $G_7$, $G_9$, $G_{10}$) examples, while get better results than GA [3], which with four better ($G_4$, $G_7$, $G_9$, $G_{10}$), two almost same ($G_6$, $G_8$) and two worse ($G_1$, $G_2$) examples.

Table 4 gives the summary of mean best results $F_B$ by the cases of DEPS. The *Boundary* mode and *Random* mode of DEPS are better than that of LPS in most examples through the complementary searching role by the hybrid DE operator. Moreover, the DEPS#P1 still performs well than the DEPS#B and the DEPS#R. By comparison the DEPS#P2 with the existing results in the Table 2, it can be seen that DEPS#P2 get better results than ES [4] and GA [3] in almost all examples.

TABLE 4. MEAN RESULTS BY THE CASES OF DEPS

| *F*. | DEPS#B ($r_f$) | DEPS#R | DEPS#P1 | DEPS#P2 |
|---|---|---|---|---|
| $G_1$ | -6.259 | -12.248 | -14.271 | -15.000 |
| $G_2$ | 0.36326 | 0.40280 | 0.48664 | 0.64330 |
| $G_4$ | -30646.43 | -30662.20 | -30665.54 | -30665.54 |
| $G_6$ | -6961.8 (74) | -6931.271 | -6961.814 | -6961.814 |
| $G_7$ | 209.300 (2) | 26.358 | 24.897 | 24.306 |
| $G_8$ | 0.095691 | 0.095425 | 0.095558 | 0.095825 |
| $G_9$ | 20819.319 | 680.690 | 680.690 | 680.630 |
| $G_{10}$ | 8378.4 (4) | 7506.5 | 7343.5 | 7049.5 |

Here the GA [3] performs the best for $G_2$ in all cases of algorithm settings. However, if the $T$ of LPS#P2 is increased to 1E4, i.e. $T_E$ is increased to 7E5, then $F_B$ of $G_2$ was 0.79298, which is better than GA [3].

*B. Engineering design examples* [10]

Table 5 gives the global minimum value $F^*$ and existing results with number of evaluations by Ray et al. [10] for engineering design examples. Here the global optimum of *SR* is located at the boundary; for *WB*, *TS*, their global optimums are closed to the boundary.

Table 6 lists the results calculated by LPS in different handling modes, where $N$=40, $T$=500, then $T_E$=2E4. For each example, 500 runs were executed. Here the solutions of *Boundary* mode also are often trapped into the local minimums at the boundary. For *Random* mode, it found worse results for the cases *WB*, *TS* and *SR*, which the global minimum are located at or closed to the boundary. Moreover, the *Periodic* mode can find better results that Ray's [10] in less evaluation times.

TABLE 5. EXISTING RESULTS FOR ENGINEERING PROBLEMS

| *Design problems* | $F^*$ | Results [10] | $T_E$ [10] |
|---|---|---|---|
| *Welded Bean* (*WB*) | 2.38113 | 2.96070 | 64862 |
| *Speed Reducer* (*SR*) | 2994.471 | 2998.027 | 110235 |
| *Three-Bar Truss* (*TB*) | 263.8958 | 263.8989 | 36113 |
| *Tension Spring (TS)* | 0.012666 | 0.012923 | 25167 |

TABLE 6. MEAN RESULTS BY LPS IN DIFFERENT MODES

| *Mode* | *WB* | *SR* | *TB* | *TS* |
|---|---|---|---|---|
| *Boundary* | 3.35408 | 3060.910 | 263.89646 | 0.013129 |
| *Random* | 2.56145 | 3100.733 | 263.89649 | 0.015372 |
| *Periodic* | 2.40403 | 2994.497 | 263.89654 | 0.012922 |

## V. CONCLUSION

This paper has analyzed a *Periodic* boundary handling mode, which is employed for improving the robustness of particle swarm. The method does not introduce any additional parameters. By providing an infinite space, which is composed of periodic copies of original $S$. it eliminates possible disorganizing for the particle swarm that caused by the unnecessary mutations at the boundary of $S$ as in conventional handling methods. Besides, it provides an effective copy of $S$ for the flying of dynamic particle swarm, which the maximum possible variation length by *Periodic* mode is also decreased to half of conventional handling methods for each dimension.

The performance of particle swarm with *Periodic* mode on benchmark functions was compared with that of conventional handling modes, include *Boundary* and *Random* mode, which produced better results, especially for the cases with local optimums that closed to or located at the boundary of $S$. It was also compared with the existing results of different algorithms, which provided better results in less evaluation times.